\documentclass[11pt,a4paper]{article}
\usepackage[hyperref]{naaclhlt2019}
\hypersetup{draft}
\usepackage{times}
\usepackage{latexsym}
\usepackage{microtype}
\usepackage{url}
\usepackage{booktabs}
\usepackage[safe]{tipa}

\aclfinalcopy 

\usepackage{xcolor}
\usepackage{soul}
\usepackage{longtable}
\usepackage{multirow}
\usepackage{booktabs}
\usepackage{amsmath}
\usepackage{amsfonts}
\usepackage[utf8x]{inputenc}
\usepackage{bm}
\usepackage{microtype}
\usepackage{tikz}
\usepackage{microtype}
\usepackage{enumitem}
\usepackage{xfrac}
\usetikzlibrary{bayesnet}
\usepackage{placeins}

\usepackage{todonotes}
\makeatletter
\newcommand*\iftodonotes{\if@todonotes@disabled\expandafter\@secondoftwo\else\expandafter\@firstoftwo\fi}  
\makeatother

\usepackage{cleveref}
\crefname{section}{\S}{\S\S}
\Crefname{section}{\S}{\S\S}
\crefname{table}{Tab.}{}
\crefname{figure}{Fig.}{}
\crefname{algorithm}{Algorithm}{}
\crefname{equation}{eq.}{}
\crefname{appendix}{App.}{}
\crefformat{section}{\S#2#1#3}  

\newcommand{\vz}{\bm{z}}

\newcommand{\saveforCR}[1]{}

\title{Combining Sentiment Lexica with a \\Multi-View Variational Autoencoder}

\date{}

\author  
  {
	\begin{tabular}{lllll}
	Alexander Hoyle\raise1.0ex\hbox{\normalfont\normalsize \textschwa}\raise1.0ex\hbox{\normalfont \normalsize} & Lawrence Wolf-Sonkin\raise1.0ex\hbox{\normalfont\normalsize \textipa{S}}\raise1.0ex\hbox{\normalfont\normalsize} & Hanna Wallach\raise1.0ex\hbox{\normalfont\normalsize \textipa{Z}} 
	\end{tabular} \\
		\begin{tabular}{lllll}
 \textbf{Ryan Cotterell}\raise1.0ex\hbox{\normalfont\normalsize \textipa{H}}
	& \textbf{Isabelle Augenstein}\raise1.0ex\hbox{\normalfont\normalsize \textipa{P}}
	\end{tabular}
	\\ 
    \raise1.0ex\hbox{\normalfont\normalsize \textschwa}University College London, London, UK \\
    \raise1.0ex\hbox{\normalfont\normalsize \textipa{S}}Department of Computer Science, Johns Hopkins University, Baltimore, USA \\
    \raise1.0ex\hbox{\normalfont\normalsize \textipa{Z}}Microsoft Research, New York City, USA \\
    \raise1.0ex\hbox{\normalfont\normalsize \textipa{H}}Department of Computer Science and Technology, University of Cambridge, Cambridge, UK\\
     \raise1.0ex\hbox{\normalfont\normalsize \textipa{P}}Department of Computer Science, University of Copenhagen, Copenhagen, Denmark\\
	{\tt \small{alexander.hoyle.17@ucl.ac.uk, lawrencews@jhu.edu}} \\ 
	{\tt \small {hanna@dirichlet.net, rdc42@cam.ac.uk, genstein@di.ku.dk}}
}

\begin{document}

\maketitle

\begin{abstract}
When assigning quantitative labels to a dataset,
different methodologies may rely on different scales. In particular, when assigning polarities to words in a sentiment lexicon, annotators may use binary, categorical, or continuous labels. Naturally,
it is of interest to unify these labels from disparate scales to both achieve maximal coverage over words and to create a single, more robust sentiment lexicon while retaining scale coherence.
We introduce a generative model of sentiment lexica to combine disparate scales into a common latent representation. We realize this model with a novel multi-view variational autoencoder (VAE), called SentiVAE.
We evaluate our approach via a downstream text classification task involving nine English-Language sentiment analysis datasets; 
our representation outperforms six individual sentiment lexica, as well as a straightforward combination thereof.\looseness=-1

\end{abstract}

\section{Introduction}

\begin{figure}
\centering
    \includegraphics[width=\columnwidth]{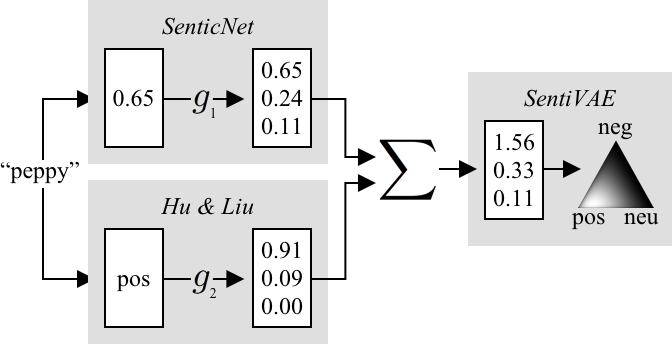}
    \caption{A depiction of the ``encoder'' portion of SentiVAE. The word \emph{peppy} has polarity values of 0.65 and \emph{pos} in the SenticNet and Hu-Liu lexica, respectively. These values are ``encoded'' into two three-dimensional vectors, which are then summed and added to $(1,1,1)$ (not shown) to form the parameters of a Dirichlet over the latent representation of the word's polarity value.}
    \label{fig:sentivae-diagram}
\end{figure}

Sentiment lexica provide an easy way to automatically label texts with polarity values, and are also frequently transformed into features for supervised models, including neural networks \cite{conf/semeval/PalogiannidiKCK16,conf/aaai/MaPC18}. 
Indeed, given their utility, a veritable cottage industry has emerged focusing on the design of sentiment lexica.
In practice, using any single lexicon, unless specifically and carefully designed for the particular domain of interest, has several downsides. For example, any lexicon will typically have low coverage compared to the language's entire vocabulary, and may have misspecified labels for the domain.
In many cases, it may therefore be desirable to combine multiple
sentiment lexica into a single representation. 
Indeed, some research on unifying such lexica has emerged \cite{conf/coling/EmersonD14,conf/amcis/AltrabshehEM17}, borrowing ideas from crowdsourcing \cite{Raykar:2010:LC:1756006.1859894,conf/naacl/HovyBVH13}. However, this is a non-trivial task, because lexica can use binary, categorical, or continuous scales to quantify polarity---in addition to different interpretations for each---and thus cannot easily be combined. In \cref{fig:sentivae-diagram}, we show an example of the same word labeled using different lexica to illustrate the nature of the challenge.\looseness=-1

\begin{figure}
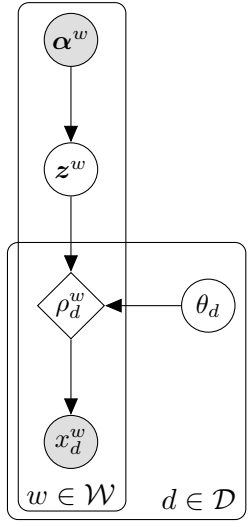

    \centering
    \tikz{ %
        \node[obs] (alpha) {$\bm{\alpha}^w$} ;
        \node[latent, below=of alpha] (z) {$\vz^w$} ;
        \node[det, below=of z] (rho) {$\rho_d^w$} ;
        \node[latent, right=of rho] (theta) {$\theta_d$} ;
        \node[obs, below=of rho] (x) {$x_d^w$} ;
        \edge {alpha} {z}; 
        \edge {z} {rho};
        \edge {rho} {x};
        \edge {theta} {rho};
        
        \node[below=of x,yshift=0.37in] (wlab) {$w\in\mathcal{W}$};
        \plate[xshift=0.09in] {W} {(alpha)(z)(x)} {\phantom{$w\in\mathcal{W}\ \ $}};
        \plate[yshift=0.1in] {D} {(theta)(x)(W.south west)(rho.north)} {$\qquad\qquad$};
        \node[below=of theta,yshift=-0.37in,xshift=-0.05in] {$d\in\mathcal{D}$};
    }
    \caption{Generative model for SentiVAE.}
    \label{fig:sentiment_vae}
\end{figure}

To combine sentiment lexica with disparate
scales, we introduce SentiVAE, a novel multi-view variant of the variational autoencoder (VAE) \cite{kingmaAutoEncodingVariationalBayes2013}. SentiVAE, visualized as a graphical model in \cref{fig:sentiment_vae}, differs from the original VAE in two
ways: (i) it uses a Dirichlet latent variable (rather than a Gaussian) for each word in the combined vocabulary, and (ii) it has multiple emission distributions---one for each lexicon. Because the latent variables are shared across the lexica,
we are able to derive a common latent representation of the words' polarities. The resulting model is spiritually related to a multi-view learning approach \cite{sun2013survey}, where each view corresponds to a different lexicon. Experimentally, we use SentiVAE to combine six commonly used English-language sentiment lexica with disparate scales.\looseness=-1

We evaluate the resulting representation via a text classification task involving nine English-language sentiment analysis datasets. For each dataset, we transform each text into an average polarity value using either our representation, one of the six commonly used sentiment lexica, or a straightforward combination thereof. We then train a classifier to predict the overall sentiment of each text from its average polarity value. We find that our representation outperforms the individual lexica, as well as the straightforward combination for some datasets. Our representation is particularly efficacious for datasets from domains that are not well-supported by standard sentiment lexica.\footnote{Our representation and code are available at \url{https://github.com/ahoho/SentiVAE}.}

The existing research that is most closely related to our work is SentiMerge \cite{conf/coling/EmersonD14}, a Bayesian approach for aligning sentiment lexica with different continuous scales. SentiMerge consists of two steps: (i) aligning the lexica via rescaling, and (ii) combining the rescaled lexica using a Gaussian distribution. The authors perform token-level evaluation using a single sentiment analysis dataset where each token is labeled with its contextually dependent sentiment. Because SentiMerge can only combine lexica with continuous scales, we do not include it in our evaluation.

\begin{table}
\centering
\resizebox{0.48\textwidth}{!}{%
\begin{tabular}{l l l c}
\toprule
Lexicon & Source & $N$ & $\text{Dom}$\\
\midrule
{\tt SentiWordNet} & WordNet & 14107 & $[-1,1]^2$  \\
{\tt MPQA} & Newswire & 4397 & $\{0, 1\}$  \\
{\tt SenticNet} & --- & 100000 & $[-1,1]$ \\
{\tt Hu-Liu} & Product reviews & 6790 & $\{0, 1\}$ \\
{\tt GI} & --- & 4206 & $\{0, 1\}$ \\
{\tt VADER} & Social media & 7489 & $\{0,\ldots,8\}^{10}$  \\
\bottomrule
\end{tabular}%
}
\caption{Descriptive statistics for the sentiment lexica. $N$: vocabulary size. $\text{Dom}$: Domain of polarity values.\looseness=-1}
\label{tab:lexica-stats}
\end{table}

\section{Sentiment Lexica and Scales}
\label{sec:lexica}
We use the following commonly used English-language sentiment lexica:
SentiWordNet \cite{baccianellaSENTIWORDNETEnhancedLexical},
MPQA \cite{wilsonRecognizingContextualPolarity},
SenticNet 5 \cite{cambriaSenticNetDiscoveringConceptual},
Hu-Liu \cite{huMiningSummarizingCustomer}, 
GI \cite{stoneGeneralInquirerComputer1962}, and
VADER \cite{huttoVADERParsimoniousRulebased}.
Descriptive statistics for each lexicon are shown in \cref{tab:lexica-stats}. Each word in SentiWordNet is labeled with two real values, each in the interval $[0, 1]$, corresponding to the strength of positive and negative sentiment (e.g., the label $(0\ 0)$ is neutral, while the label $(1\ 0)$ is maximally positive). Each word in VADER is labeled by ten different human evaluators, with each evaluator providing a polarity value on a nine-point scale (where the midpoint is neutral), yielding a 10-dimensional label. MPQA, Hu-Liu, and GI all use binary scales. Lastly, each word in SenticNet is labeled with a real value in the interval $[-1,1]$, where 0 is neutral. 

\section{SentiVAE}
We first describe a figurative generative process for a single sentiment lexicon $d \in \mathcal{D}$, where $\mathcal{D}$ is a set of sentiment lexica. Imagine there is a true (latent) polarity value $\bm{z}^w$ associated with each word $w$ in the lexicon's vocabulary. When the lexicon's creator labels that word according to their chosen scale (e.g., thumbs-up or thumbs-down, a real value in the interval $[0, 1]$), they deterministically transform this true value to their chosen scale via a function $f(\,\cdot\,;\,\bm{\theta}_d)$.\footnote{Parameterized by lexicon-specific weights $\bm{\theta}_d$.} Sometimes, noise is introduced during this labeling process, corrupting the label as it leaves the ethereal realm and producing the (observed) polarity label $\bm{x}^w_d$. They then add this potentially noisy label to the lexicon.\looseness=-1

Given a lexicon of observed polarity labels, the latent polarity values can be inferred using a VAE. The original VAE posits a generative model of observed data $\mathcal{X}$ and latent variables $\mathcal{Z}$: $P(\mathcal{X}, \mathcal{Z}) = P(\mathcal{X} \mid \mathcal{Z})\,P(\mathcal{Z})$. Inference of $\mathcal{Z}$ then proceeds by approximating the (intractable) posterior $P(\mathcal{Z} \mid \mathcal{X})$ with a Gaussian distribution, factorized over the individual latent variables. A parameterized encoder function compresses $\mathcal{X}$ into $\mathcal{Z}$, while a parameterized decoder function reconstructs $\mathcal{X}$ from $\mathcal{Z}$.

SentiVAE extends the original VAE model to combine multiple lexica with disparate scales, producing a common latent representation of the polarity value for each word in the combined vocabulary.

\paragraph{Generative process.}
Given a set of sentiment lexica $\mathcal{D}$ with a combined vocabulary $\mathcal{W}$, SentiVAE posits a common latent representation $\vz^w$ of the polarity value for each word $w \in \mathcal{W}$, where $\vz^w$ is a three-dimensional categorical distribution over the sentiments \textit{positive}, \textit{negative}, and \textit{neutral}.

The generative process starts by drawing each latent polarity value $\vz^w$ from a three-dimensional Dirichlet prior, parameterized by $\bm{\alpha}^w = (1,1,1)$:\looseness=-1
\begin{equation}
	\vz^w \sim \text{Dir}(\bm{\alpha}^w).
\end{equation}
If the word is uncontroversial,\footnote{We say that a word is uncontroversial if there is strong agreement across the sentiment lexica in which it appears. Even without this spurring, the inferred latent representation typically separates into the three sentiment classes, but performance on our text classification task is somewhat diminished.\looseness=-1} we spur this prior somewhat using the number of lexica in which the word appears $c(w)$. Specifically, we add $c(w)$ to the parameter for the sentiment associated with that word in the lexica, e.g., $\bm{\alpha}^\textsc{superb} = (1 + c(\textsc{superb}), 1, 1)$. This has the effect of regularizing the inferred latent polarity value toward the desired distribution over sentiments.

Having generated $\vz^w$, the process proceeds by ``decoding'' $\vz^w$ into each lexicon's chosen scale. First, for each lexicon $d \in \mathcal{D}$, $\vz^w$ is deterministically transformed via neural network $f(\,\cdot\,;\,\bm{\theta}_d)$ with a single 32-dimensional hidden layer, parameterized by lexicon-specific weights $\bm{\theta}_d$:\looseness=-1
\begin{equation}
  \bm{\rho}_d^w = f(\bm{z}^w; \bm{\theta}_d).
\end{equation}
The transformed value $\bm{\rho}_d^w$ is then used to generate the (observed)
polarity label $\bm{x}^w_d$ for that lexicon:
\begin{equation}
  \bm{x}_d^w \sim P_d(\bm{x}_d^w \mid \bm{\rho}_d^w).
\end{equation}
The dimensionality of $\bm{\rho}_d^w$ and the emission distribution
$P_d$ are lexicon-specific. For SentiWordNet, $P_d$ is a
two-dimensional Gaussian with mean $\bm{\rho}_d^w$ and a diagonal
  covariance matrix equal to $0.01 \bm{I}$; for VADER, $P_d$ consists of ten nine-dimensional categorical distributions, collectively parameterized by $\bm{\rho}_d^w$; for MPQA, Hu-Liu, and
  GI, $P_d$ is a Bernoulli distribution, parameterized by $\bm{\rho}_d^w$; and for SenticNet, $P_d$ is a univariate Gaussian with mean and variance each an element in a two-dimensional $\bm{\rho}_d^w$.\looseness=-1 

\begin{table}
\centering
\resizebox{0.48\textwidth}{!}{%
\begin{tabular}{l l l l l}
\toprule
Dataset & Source & $N$ & Classes\\
\midrule
{\tt IMDB} & Movies & 25000 & 2 \\
{\tt Yelp} & Product reviews & 100000 & 5 / 3 \\
{\tt SemEval} & Twitter & 7668 & 3 \\
{\tt MultiDom} & Product reviews & 6500 & 2 \\
{\tt ACL} & Scientific reviews & 248 & 5 / 3 \\
{\tt ICLR} & Scientific reviews & 2166 & 10 / 3 \\
\bottomrule
\end{tabular}%
}
\caption{Descriptive statistics for the training portions of the sentiment analysis datasets. $N$: number of texts.}
\label{tab:dataset-stats}
\end{table}

 \paragraph{Inference.} Inference involves forming the posterior distribution over the latent polarity values $\mathcal{Z}$ given the observed polarity labels $\mathcal{X}$. Because computing the normalizing constant $P(\mathcal{X})$ is intractable, we instead approximate the posterior with a family of distributions $Q_{\bm{\lambda}}(\mathcal{Z})$, indexed by variational parameters $\bm{\lambda}$. Specifically, we use
  \begin{equation}
    Q_{\bm{\lambda}}(\mathcal{Z}) = \prod_{w \in \mathcal{W}} Q_{\bm{\beta}^w}(\bm{z}^w) = \prod_{w \in \mathcal{W}} \text{Dir}(\bm{\beta}^w).
  \end{equation}
  To construct $\bm{\beta}^w$, we first define a neural network $g(\cdot;\, \bm{\phi}_d)$, with a single 32-dimensional hidden layer, which ``encodes'' $\bm{x}_d^w$ into a three-dimensional vector. The output of this neural network is then transformed via a softmax as follows:
\begin{align}
        \bm{\omega}^w_d &= \text{softmax}\big( g(\bm{x}_d^w ; \,\phi_d ) \big) \\
    \bm{\beta}^w &= 1 + \sum_{d \in \mathcal{D}} \bm{\omega}^w_d.
\end{align}
The intuition behind $\bm{\beta}_w$ can be understood by appealing to the ``pseudocount'' interpretation of Dirichlet parameters. Each lexicon contributes exactly one pseudocount, divided among \emph{positive}, \emph{negative}, and \emph{neutral}, to what would otherwise be a symmetric, uniform Dirichlet distribution. As a consequence of this construction, words that appear in more lexica will have more concentrated Dirichlets. Intuitively, this property is appealing.

\begin{table*}
\centering
\resizebox{1.0\textwidth}{!}{%
\begin{tabular}{r c c c c c c c c c}
\toprule
\bf  & \bf IMDB 2C & \bf Yelp 5C & \bf Yelp 3C & \bf SemEval 3C & \bf MultiDom 2C & \bf ACL 5C & \bf ACL 3C & \bf ICLR 10C & \bf ICLR 3C \\ 
\midrule
\textbf{SentiVAE}\quad$\text{E}_Q[\vz^w]$
                              &     72.7     &     49.8     &     57.5     &     46.0     &     70.8     &     66.7     &     73.3     &     92.6     & \bf 87.0          \\
\textbf{SentiVAE}\quad\makebox[0pt][l]{$\bm{\beta}^w$}\phantom{$\text{E}[\vz]$}     
                              &     73.4     &     49.7     &     59.4     & \bf 52.2     &     74.7     & \bf 73.3     & \bf 80.0     &     92.6     &     86.5          \\
 \midrule                                                                                                    
\bf SentiWordNet              &     63.4     &     36.0     &     47.6     &     32.2     &     62.0     &     60.0     &     53.3     &     89.1     &     83.5          \\
\bf MPQA                      &     65.4     &     44.0     &     53.0     &     29.9     &     67.4     &     60.0     &     53.3     &     89.1     &     83.5          \\
\bf SenticNet                 &     60.5     &     38.4     &     43.4     &     37.2     &     62.3     &     60.0     &     53.3     &     89.1     &     83.9          \\
\bf Hu-Liu                      &     67.2     &     46.6     &     56.4     &     31.5     &     69.4     &     60.0     &     53.3     &     89.1     &     83.5          \\
\bf GI                        &     58.4     &     40.7     &     47.9     &     31.3     &     61.6     &     60.0     &     53.3     &     89.1     &     83.5          \\
\bf VADER                     &     71.7     &     46.8     &     59.3     &     38.5     &     73.5     &     66.7     &     66.7     & \bf 94.3     &     86.1          \\
\midrule                                                                                                    
\bf Combined                  & \bf 75.6     & \bf 51.0     & \bf 64.1     &     50.6     & \bf 75.4     &     66.7     &     66.7     &     93.9     &     86.1          \\
\bottomrule
\end{tabular}%
}
\caption{Classification accuracies for our representation, six lexica, and a straightforward combination thereof.}
\label{tab:results_en}
\end{table*}

We optimize the resulting ELBO objective \cite{Blei2016VariationalIA} with respect to the variational parameters via stochastic variational inference \cite{journals/jmlr/HoffmanBWP13} using Adam \cite{Kingma2014AdamAM} in the Pyro framework \cite{journals/corr/abs-1810-09538}. The standard reparameterization trick used in the original VAE does not apply to models with Dirichlet-distributed latent variables, so we use the generalized reparameterization trick of \citet{conf/nips/RuizTB16}.

\section{Experiments and Results}
To evaluate our approach, we first use SentiVAE to combine the six
lexica described in \cref{sec:lexica}. For each word $w$ in the combined vocabulary, we obtain an estimate of $\bm{z}^w$ by taking the mean of $Q_{\bm{\beta}^w}(\bm{z}^w) = \text{Dir}(\bm{\beta}^w)$---i.e., by normalizing $\bm{\beta}^w$. We compare this representation to using $\bm{\beta}^w$ directly, because $\bm{\beta}^w$ contains information about SentiVAE's certainty about the word's latent polarity value. We evaluate our common latent 
representation via a text classification task involving nine
English-language sentiment analysis datasets: IMDB
\cite{maas-EtAl:2011:ACL-HLT2011}, Yelp
\cite{zhang2015characterlevel}, SemEval 2017 Task 4 (SemEval,
\citet{SemEval:2017:task4}), multi-domain sentiment analysis
(MultiDom, \citet{blitzer-dredze-pereira:2007:ACLMain}), and PeerRead
\cite{conf/naacl/KangADZKHS18} with splits ACL 2017 and ICLR 2017
\cite{conf/naacl/KangADZKHS18}. Each dataset consists of multiple
texts (e.g., tweets, articles), each labeled with an overall
sentiment (e.g., \emph{positive}). Descriptive statistics for each
dataset are shown in \cref{tab:dataset-stats}. For the datasets with more than three sentiment labels, we consider two versions---the original and a version with only three (bucketed) sentiment labels.

For each dataset, we transform each text into an average polarity
value using either our representation, one of the six lexica,\footnote{We bucket the upper four and lower four points of VADER's nine-point scale, to yield a three-point scale. Without this bucketing, our representation outperforms VADER on four of the nine datasets. We do not bucket VADER when using it in SentiVAE or in the straightforward combination.} or a straightforward combination thereof, where the polarity value for each
word in the (combined) vocabulary is a 16-dimensional
vector that consists of a concatenation of polarity values. (Unlike SentiVAE, this concatenation does not yield a single sentiment lexicon that retains scale coherence, while achieving 
maximal coverage over words.) Specifically, we replace each token with its corresponding
polarity value, and then average the these values
\cite{Go_Bhayani_Huang_2009,conf/ranlp/OzdemirB15,journals/jair/KiritchenkoZM14}. We
then use the training portion of the dataset to learn a logistic regression classifier to predict the overall sentiment of each text
from its average polarity value. Finally, we use the testing portion to compute the accuracy of the classifier.


\paragraph{Results.}
The results in \cref{tab:results_en} show that our representation
using $\bm{\beta}^w$ outperforms the individual lexica for all but one
dataset, and that our representation using the mean of
$Q_{\bm{\beta}^w}(\vz^w)$ outperforms them for six datasets. This is
likely because SentiVAE has a richer representation of sentiment than
any individual lexicon, and it has greater coverage over words (see
\cref{tab:lexica-coverage}). The results in
\cref{tab:results-intersection} support the former reason: even when
we limit the words in our representation to match those in an
individual lexicon, our representation still outperforms the
individual lexicon. Unsurprisingly, our representation
especially outperforms lexica with unidimensional scales. We also find
that our representation outperforms the straightforward combination
for datasets from domains that are not well supported by the
individual lexica (see Tabs.~\cref{tab:lexica-stats,tab:dataset-stats}
for lexicon and dataset sources, respectively). By combining 
lexica from different domains, our representation captures a general
notion of sentiment that is not tailored to any specific domain.

\begin{table}[t]
\footnotesize
\begin{tabular}{r c c c c}
\toprule
                        & \bf IMDB      & \bf SemEval & \bf Multi & \bf ICLR \\
    \midrule
    {\bf SentiVAE}       & 70       & 64  & 81  &  71  \\
    \midrule
    {\bf SentiWordNet}   & 15       & 14  & 24  &  16  \\
    {\bf MPQA}           & 10       & 7   & 18  &  9   \\
    {\bf SenticNet}      & 40       & 39  & 53  &  45  \\
    {\bf Hu-Liu}         & 7        & 5   & 13  &  5   \\
    {\bf GI}             & 8        & 7   & 15  &  6   \\
    {\bf VADER}          & 7        & 6   & 13  &  5   \\
\bottomrule
\end{tabular}
\caption{Coverage over words (percentage) by lexicon for the training portions of four of the nine datasets.}
\label{tab:lexica-coverage}
\end{table}

\section{Conclusion}

We introduced a generative model of sentiment lexica to combine
disparate scales into a common latent representation, and realized
this model with a novel multi-view variational autoencoder, called
SentiVAE. We then used SentiVAE to combine six commonly used
English-language sentiment lexica with binary, categorical, and
continuous scales. Via a downstream text classification task involving
nine English-language sentiment analysis datasets, we found that our
representation outperforms the individual lexica, as well as a
straightforward combination thereof. We also found that our
representation is particularly efficacious for datasets from domains
that are not well-supported by standard sentiment lexica. Finally, we
note that our approach is more general than SentiMerge
\cite{conf/coling/EmersonD14}. While SentiMerge can only combine
sentiment lexica with continuous scales, SentiVAE is designed to
combine lexica with disparate scales.

\section{Acknowledgements}
We would like to thank to Adam Forbes for the design of \cref{fig:sentivae-diagram}.
We further acknowledge the support of the NVIDIA Corporation with the donation of the Titan Xp GPU used to conduct this research.

\begin{table}[t]
\centering
\footnotesize
\begin{tabular}{r c c c c c}
\toprule
    & \multicolumn{2}{c}{\bf IMDB 2C} & & \multicolumn{2}{c}{\bf SemEval 3C} \\
    \midrule
                       & SV & Lex  & & SV  & Lex  \\
    \midrule
    \bf SentiVAE       &\bf74.7 & --    & &\bf72.4 &  --  \\
    \midrule
    \bf SentiWordNet   &   70.6 & 63.4 & &   67.4 & 55.1 \\
    \bf MPQA           &   73.5 & 66.6 & &   62.6 & 51.8 \\
    \bf SenticNet      &   74.4 & 60.9 & &   72.1 & 59.5 \\
    \bf Hu-Liu         &   73.6 & 68.4 & &   59.1 & 51.1 \\
    \bf GI             &   71.4 & 59.3 & &   63.8 & 54.0 \\
    \bf VADER          &   73.6 & 73.1 & &   60.9 & 58.7 \\
\bottomrule
\end{tabular}
\caption{Classification accuracies for a 10\% validation portion of two of the datasets. The first row, labeled SentiVAE, contains the classification accuracy for our representation using $\bm{\beta}^w$. Subsequent (lexicon-specific) rows compare our representation (SV), restricted to the vocabulary of that lexicon, to the lexicon itself (Lex).}
\label{tab:results-intersection}
\end{table}

\bibliography{naaclhlt2019}

\begin{thebibliography}{27}
\expandafter\ifx\csname natexlab\endcsname\relax\def\natexlab#1{#1}\fi

\bibitem[{Altrabsheh et~al.(2017)Altrabsheh, El-Masri, and
  Mansour}]{conf/amcis/AltrabshehEM17}
Nabeela Altrabsheh, Mazen El-Masri, and Hanady Mansour. 2017.
\newblock {Combining Sentiment Lexicons of Arabic Terms}.
\newblock In \emph{AMCIS}. Association for Information Systems.

\bibitem[{Baccianella et~al.(2010)Baccianella, Esuli, and
  Sebastiani}]{baccianellaSENTIWORDNETEnhancedLexical}
Stefano Baccianella, Andrea Esuli, and Fabrizio Sebastiani. 2010.
\newblock {SentiWordNet 3.0: An Enhanced Lexical Resource for Sentiment
  Analysis and Opinion Mining}.
\newblock 10(2010):2200--2204.

\bibitem[{Bingham et~al.(2018)Bingham, Chen, Jankowiak, Obermeyer, Pradhan,
  Karaletsos, Singh, Szerlip, Horsfall, and
  Goodman}]{journals/corr/abs-1810-09538}
Eli Bingham, Jonathan~P. Chen, Martin Jankowiak, Fritz Obermeyer, Neeraj
  Pradhan, Theofanis Karaletsos, Rohit Singh, Paul~A. Szerlip, Paul Horsfall,
  and Noah~D. Goodman. 2018.
\newblock {Pyro: Deep Universal Probabilistic Programming}.
\newblock \emph{CoRR}, abs/1810.09538.

\bibitem[{Blei et~al.(2017)Blei, Kucukelbir, and
  McAuliffe}]{Blei2016VariationalIA}
David~M. Blei, Alp Kucukelbir, and Jon~D. McAuliffe. 2017.
\newblock Variational inference: {A} review for statisticians.
\newblock \emph{Journal of the American Statistical Association}, 112:859--877.

\bibitem[{Blitzer et~al.(2007)Blitzer, Dredze, and
  Pereira}]{blitzer-dredze-pereira:2007:ACLMain}
John Blitzer, Mark Dredze, and Fernando Pereira. 2007.
\newblock {Biographies, Bollywood, Boom-boxes and Blenders: Domain Adaptation
  for Sentiment Classification}.
\newblock In \emph{Proceedings of the 45th Annual Meeting of the Association of
  Computational Linguistics}, pages 440--447, Prague, Czech Republic.
  Association for Computational Linguistics.

\bibitem[{Cambria et~al.(2014)Cambria, Olsher, and
  Rajagopal}]{cambriaSenticNetDiscoveringConceptual}
Erik Cambria, Daniel Olsher, and Dheeraj Rajagopal. 2014.
\newblock {SenticNet 3: A Common and Common-sense Knowledge Base for
  Cognition-driven Sentiment Analysis}.
\newblock In \emph{Proceedings of the Twenty-Eighth AAAI Conference on
  Artificial Intelligence}, AAAI'14, pages 1515--1521. AAAI Press.

\bibitem[{Emerson and Declerck(2014)}]{conf/coling/EmersonD14}
Guy Emerson and Thierry Declerck. 2014.
\newblock {SentiMerge: Combining Sentiment Lexicons in a Bayesian Framework}.
\newblock In \emph{Proceedings of Workshop on Lexical and Grammatical Resources
  for Language Processing}, pages 30--38. Association for Computational
  Linguistics and Dublin City University.

\bibitem[{Go et~al.(2009)Go, Bhayani, and Huang}]{Go_Bhayani_Huang_2009}
Alec Go, Richa Bhayani, and Lei Huang. 2009.
\newblock {Twitter Sentiment Classification using Distant Supervision}.
\newblock \emph{Processing}, pages 1--6.

\bibitem[{Hoffman et~al.(2013)Hoffman, Blei, Wang, and
  Paisley}]{journals/jmlr/HoffmanBWP13}
Matthew~D. Hoffman, David~M. Blei, Chong Wang, and John~William Paisley. 2013.
\newblock Stochastic variational inference.
\newblock \emph{Journal of Machine Learning Research}, 14(1):1303--1347.

\bibitem[{Hovy et~al.(2013)Hovy, Berg-Kirkpatrick, Vaswani, and
  Hovy}]{conf/naacl/HovyBVH13}
Dirk Hovy, Taylor Berg-Kirkpatrick, Ashish Vaswani, and Eduard~H. Hovy. 2013.
\newblock {Learning Whom to Trust with MACE}.
\newblock In \emph{Proceedings of the 2013 Conference of the North American
  Chapter of the Association for Computational Linguistics: Human Language
  Technologies}, pages 1120--1130. Association for Computational Linguistics.

\bibitem[{Hu and Liu(2004)}]{huMiningSummarizingCustomer}
Minqing Hu and Bing Liu. 2004.
\newblock Mining and summarizing customer reviews.
\newblock In \emph{Proceedings of the Tenth ACM SIGKDD International Conference
  on Knowledge Discovery and Data Mining}, pages 168--177. ACM.

\bibitem[{Hutto and Gilbert(2014)}]{huttoVADERParsimoniousRulebased}
C.~J. Hutto and Eric Gilbert. 2014.
\newblock {VADER: A Parsimonious Rule-Based Model for Sentiment Analysis of
  Social Media Text}.
\newblock In \emph{Eighth International Conference on Weblogs and Social Media
  (ICWSM-14)}.

\bibitem[{Kang et~al.(2018)Kang, Ammar, Dalvi, van Zuylen, Kohlmeier, Hovy, and
  Schwartz}]{conf/naacl/KangADZKHS18}
Dongyeop Kang, Waleed Ammar, Bhavana Dalvi, Madeleine van Zuylen, Sebastian
  Kohlmeier, Eduard~H. Hovy, and Roy Schwartz. 2018.
\newblock {A Dataset of Peer Reviews (PeerRead): Collection, Insights and NLP
  Applications}.
\newblock In \emph{Proceedings of the 2018 Conference of the North American
  Chapter of the Association for Computational Linguistics: Human Language
  Technologies, Volume 1 (Long Papers)}, pages 1647--1661. Association for
  Computational Linguistics.

\bibitem[{Kingma and Ba(2015)}]{Kingma2014AdamAM}
Diederik~P. Kingma and Jimmy Ba. 2015.
\newblock Adam: {A} method for stochastic optimization.
\newblock In \emph{International Conference on Learning Representations
  (ICLR)}.

\bibitem[{Kingma and Welling(2014)}]{kingmaAutoEncodingVariationalBayes2013}
Diederik~P. Kingma and Max Welling. 2014.
\newblock {Auto-Encoding Variational Bayes}.
\newblock In \emph{Proceedings of the Second International Conference on
  Learning Representations (ICLR)}.

\bibitem[{Kiritchenko et~al.(2014)Kiritchenko, Zhu, and
  Mohammad}]{journals/jair/KiritchenkoZM14}
Svetlana Kiritchenko, Xiaodan Zhu, and Saif~M. Mohammad. 2014.
\newblock {Sentiment Analysis of Short Informal Texts}.
\newblock \emph{Journal of Machine Learning Research}, 50:723--762.

\bibitem[{Ma et~al.(2018)Ma, Peng, and Cambria}]{conf/aaai/MaPC18}
Yukun Ma, Haiyun Peng, and Erik Cambria. 2018.
\newblock \href
  {https://www.aaai.org/ocs/index.php/AAAI/AAAI18/schedConf/presentations}
  {{Targeted Aspect-Based Sentiment Analysis via Embedding Commonsense
  Knowledge into an Attentive LSTM}}.
\newblock In \emph{AAAI}, pages 5876--5883. AAAI Press.

\bibitem[{Maas et~al.(2011)Maas, Daly, Pham, Huang, Ng, and
  Potts}]{maas-EtAl:2011:ACL-HLT2011}
Andrew~L. Maas, Raymond~E. Daly, Peter~T. Pham, Dan Huang, Andrew~Y. Ng, and
  Christopher Potts. 2011.
\newblock {Learning Word Vectors for Sentiment Analysis}.
\newblock In \emph{Proceedings of the 49th Annual Meeting of the Association
  for Computational Linguistics: Human Language Technologies}, pages 142--150,
  Portland, Oregon, USA. Association for Computational Linguistics.

\bibitem[{{\"O}zdemir and Bergler(2015)}]{conf/ranlp/OzdemirB15}
Canberk {\"O}zdemir and Sabine Bergler. 2015.
\newblock {A Comparative Study of Different Sentiment Lexica for Sentiment
  Analysis of Tweets}.
\newblock In \emph{Proceedings of the International Conference Recent Advances
  in Natural Language Processing}, pages 488--496. INCOMA Ltd. Shoumen,
  BULGARIA.

\bibitem[{Palogiannidi et~al.(2016)Palogiannidi, Kolovou, Christopoulou,
  Kokkinos, Iosif, Malandrakis, Papageorgiou, Narayanan, and
  Potamianos}]{conf/semeval/PalogiannidiKCK16}
Elisavet Palogiannidi, Athanasia Kolovou, Fenia Christopoulou, Filippos
  Kokkinos, Elias Iosif, Nikolaos Malandrakis, Haris Papageorgiou, Shrikanth
  Narayanan, and Alexandros Potamianos. 2016.
\newblock {Tweester at SemEval-2016 Task 4: Sentiment Analysis in Twitter Using
  Semantic-Affective Model Adaptation}.
\newblock In \emph{Proceedings of the 10th International Workshop on Semantic
  Evaluation (SemEval-2016)}, pages 155--163. The Association for Computer
  Linguistics.

\bibitem[{Raykar et~al.(2010)Raykar, Yu, Zhao, Valadez, Florin, Bogoni, and
  Moy}]{Raykar:2010:LC:1756006.1859894}
Vikas~C. Raykar, Shipeng Yu, Linda~H. Zhao, Gerardo~Hermosillo Valadez, Charles
  Florin, Luca Bogoni, and Linda Moy. 2010.
\newblock {Learning From Crowds}.
\newblock \emph{Journal of Machine Learning Research}, 11:1297--1322.

\bibitem[{Rosenthal et~al.(2017)Rosenthal, Farra, and
  Nakov}]{SemEval:2017:task4}
Sara Rosenthal, Noura Farra, and Preslav Nakov. 2017.
\newblock {SemEval-2017 Task 4: Sentiment Analysis in Twitter}.
\newblock In \emph{Proceedings of the 11th International Workshop on Semantic
  Evaluation (SemEval-2017)}, pages 502--518, Vancouver, Canada. Association
  for Computational Linguistics.

\bibitem[{Ruiz et~al.(2016)Ruiz, Titsias, and Blei}]{conf/nips/RuizTB16}
Francisco~R. Ruiz, Michalis~K. Titsias, and David~M. Blei. 2016.
\newblock \href
  {http://papers.nips.cc/book/advances-in-neural-information-processing-systems-29-2016}
  {{The Generalized Reparameterization Gradient}}.
\newblock In \emph{Advances in Neural Information Processing Systems}, pages
  460--468.

\bibitem[{Stone et~al.(1962)Stone, Bales, Namenwirth, and
  Ogilvie}]{stoneGeneralInquirerComputer1962}
Philip~J. Stone, Robert~F. Bales, J.~Zvi Namenwirth, and Daniel~M. Ogilvie.
  1962.
\newblock The general inquirer: {A} computer system for content analysis and
  retrieval based on the sentence as a unit of information.
\newblock \emph{Behavioral Science}, 7(4):484--498.

\bibitem[{Sun(2013)}]{sun2013survey}
Shiliang Sun. 2013.
\newblock {A Survey on Multi-view Learning}.
\newblock \emph{Neural Computing and Applications}, 23(7-8):2031--2038.

\bibitem[{Wilson et~al.(2005)Wilson, Wiebe, and
  Hoffmann}]{wilsonRecognizingContextualPolarity}
Theresa Wilson, Janyce Wiebe, and Paul Hoffmann. 2005.
\newblock {Recognizing Contextual Polarity in Phrase-Level Sentiment Analysis}.
\newblock In \emph{Proceedings of Human Language Technology Conference and
  Conference on Empirical Methods in Natural Language Processing}, pages
  347--354, Vancouver, British Columbia, Canada. Association for Computational
  Linguistics.

\bibitem[{Zhang et~al.(2015)Zhang, Zhao, and LeCun}]{zhang2015characterlevel}
Xiang Zhang, Junbo Zhao, and Yann LeCun. 2015.
\newblock \href
  {http://papers.nips.cc/book/advances-in-neural-information-processing-systems-28-2015}
  {{Character-level Convolutional Networks for Text Classification}}.
\newblock In \emph{Advances in Neural Information Processing Systems}, pages
  649--657.

\end{thebibliography}
\bibliographystyle{acl_natbib}


\end{document}